 \documentclass[pmlr,twocolumn]{jmlr} 

\usepackage{booktabs}
\usepackage[load-configurations=version-1]{siunitx} 

\theorembodyfont{\upshape}
\theoremheaderfont{\scshape}
\theorempostheader{:}
\theoremsep{\newline}

\jmlrvolume{ML4H Extended Abstract Arxiv Index}
\jmlryear{2020}
\jmlrsubmitted{2020}
\jmlrpublished{}
\jmlrworkshop{Machine Learning for Health (ML4H) 2020}

\usepackage{enumitem}
\setlist[itemize]{align=parleft,left=0em..1.5em}
\setlist[enumerate]{align=parleft,left=0em..1.5em}
\usepackage[utf8]{inputenc} 
\usepackage[T1]{fontenc}    
\usepackage{booktabs}       
\usepackage{amsfonts}       
\usepackage{nicefrac}       
\usepackage{microtype}      
\usepackage{commath}
\usepackage{longtable}
\usepackage[graphicx]{realboxes}
\usepackage{placeins}
\usepackage[font={small}]{caption}
\usepackage{textcomp}
\usepackage{dcolumn}
\usepackage{mathtools}
\usepackage{comment}
\usepackage{algorithm}
\usepackage{stmaryrd}
\usepackage{algorithmic}
\usepackage[absolute,overlay]{textpos}
\usepackage{float}
\usepackage[load-configurations=version-1]{siunitx}

\newsavebox\CBox
\def\textBF#1{\sbox\CBox{#1}\resizebox{\wd\CBox}{\ht\CBox}{\textbf{#1}}}

\definecolor{lightblue}{rgb}{0.01, 0.6, 1.0}

\title[LoS Prediction in the ICU with TPC Networks]{Predicting Length of Stay in the Intensive Care Unit with \\Temporal Pointwise Convolutional Networks}

\author{
 \Name{Emma Rocheteau} 
 \Email{ecr38@cam.ac.uk}\\
 \addr University of Cambridge 
 \AND
 \Name{Pietro Li\`{o}} 
 \Email{pl219@cam.ac.uk}\\
 \addr University of Cambridge
 \AND
 \Name{Stephanie Hyland}
 \Email{stephanie.hyland@microsoft.com}\\
 \addr Microsoft Research
 }

\begin{document}

\begin{textblock}{20}(2.3,15.5)
\Large{\textcolor{gray}{This is a condensed version of \url{https://arxiv.org/abs/2007.09483}}}
\end{textblock}

\maketitle

\begin{abstract}
    We present a new deep learning architecture based on the combination of temporal convolution and pointwise (1x1) convolution to predict length of stay on the eICU critical care dataset. The model -- which we refer to as Temporal Pointwise Convolution (TPC) -- is specifically designed to mitigate for common challenges with Electronic Health Records (EHRs), such as skewness, irregular sampling and missing data. In doing so, we have achieved significant performance benefits of 18-51\% (metric dependent) over the commonly used Long-Short Term Memory (LSTM) network, and the Transformer, a multi-head self-attention network.
\end{abstract}

\section{Introduction}
In-patient length of stay (LoS) is strongly associated with inflated hospital costs \citep{rapoport2003} and the risk of hospital acquired infection and mortality \citep{hassan, LAUPLAND2006954}. Improved hospital bed planning has the potential to reduce these risks \citep{Blom2015ThePO}. However, this relies on accurate discharge date estimates\footnote{Currently, these are usually done manually by clinicians meaning they rapidly become out-of-date \citep{Nassar2016ICUPA} and can be unreliable (\citet{Mak2012PhysiciansAT} finds the mean clinician error to be 3.82 $\pm$ 6.51 days).}.

We propose a new deep learning architecture to improve automatic LoS prediction. Like in \citet{harutyunyan}, we predict the remaining LoS of patients in the ICU. Our model combines the strengths of Temporal Convolutional Layers \citep{Simonyan2016, DBLP:journals/corr/KalchbrennerESO16} in capturing causal dependencies across time, and Pointwise Convolutional Layers \citep{lin2013network} in computing higher level features from interactions in the feature domain. We show that these methods complement each other by extracting different information. Our model outperforms the commonly used Long-Short Term Memory (LSTM) network~\citep{10.1162/neco.1997.9.8.1735} and the Transformer~\citep{46201}.

We also make a case for using the mean squared logarithmic error (MSLE) loss function to train LoS models, as it helps to mitigate for positive skew in the LoS. Our code is available at: \url{https://github.com/EmmaRocheteau/eICU-LoS-prediction}.

\section{Related Work}

LSTMs have been by far the most popular model for predicting LoS and have achieved state-of-the-art results \citep{harutyunyan,sheikhalishahi2019benchmarking,Rajkomar2018ScalableAA}. They have also been applied to other patient prediction tasks e.g.\ forecasting diagnoses and medications \citep{Choi2015DoctorAP,Lipton2015LearningTD}, and mortality prediction \citep{Che2018,harutyunyan,Shickel2019DeepSOFAAC}. More recently, the Transformer model \citep{46201} -- which was originally designed for natural language processing (NLP) -- has marginally outperformed the LSTM on the LoS task \citep{2304ed73e858419398e3ee1508af5825}. Therefore, the LSTM and the Transformer were chosen as key baselines. 

\section{Methods}
We design our model to extract both trends and inter-feature relationships. Clinicians do this when assessing their patients e.g.\ they might check how the respiratory rate is changing over time, and they may also look at combination features e.g.\ the PaO$_2$/FiO$_2$ ratio.

\paragraph{Temporal Convolution}
Temporal Convolution Networks (TCNs) \citep{Simonyan2016, DBLP:journals/corr/KalchbrennerESO16} are models that convolve over the time dimension. We use stacked TCNs to extract \textit{temporal trends} in our data. Unlike most implementations, we \textit{do not share weights across features} i.e.\ weight sharing is only across time (like in Xception \citep{DBLP:journals/corr/Chollet16a}). This is because our features differ sufficiently in their temporal characteristics to warrant specialised processing. In TCNs, the receptive field sizes\footnote{`Receptive field' refers to the width of the filter. For TCNs this corresponds to a timespan.} are highly adaptable. They can be increased by using greater dilation, larger kernel sizes or by stacking more layers. By contrast, RNNs can only process one time step at a time. 

\paragraph{Pointwise Convolution}
Pointwise (or 1x1) convolution \citep{lin2013network} is typically used to reduce the dimensions in an input \citep{DBLP:journals/corr/SzegedyLJSRAEVR14}. However we use it to compute \textit{interaction features} from the existing feature set at each timepoint.

\begin{figure}[!ht]
  \centering
  \includegraphics[scale=0.38]{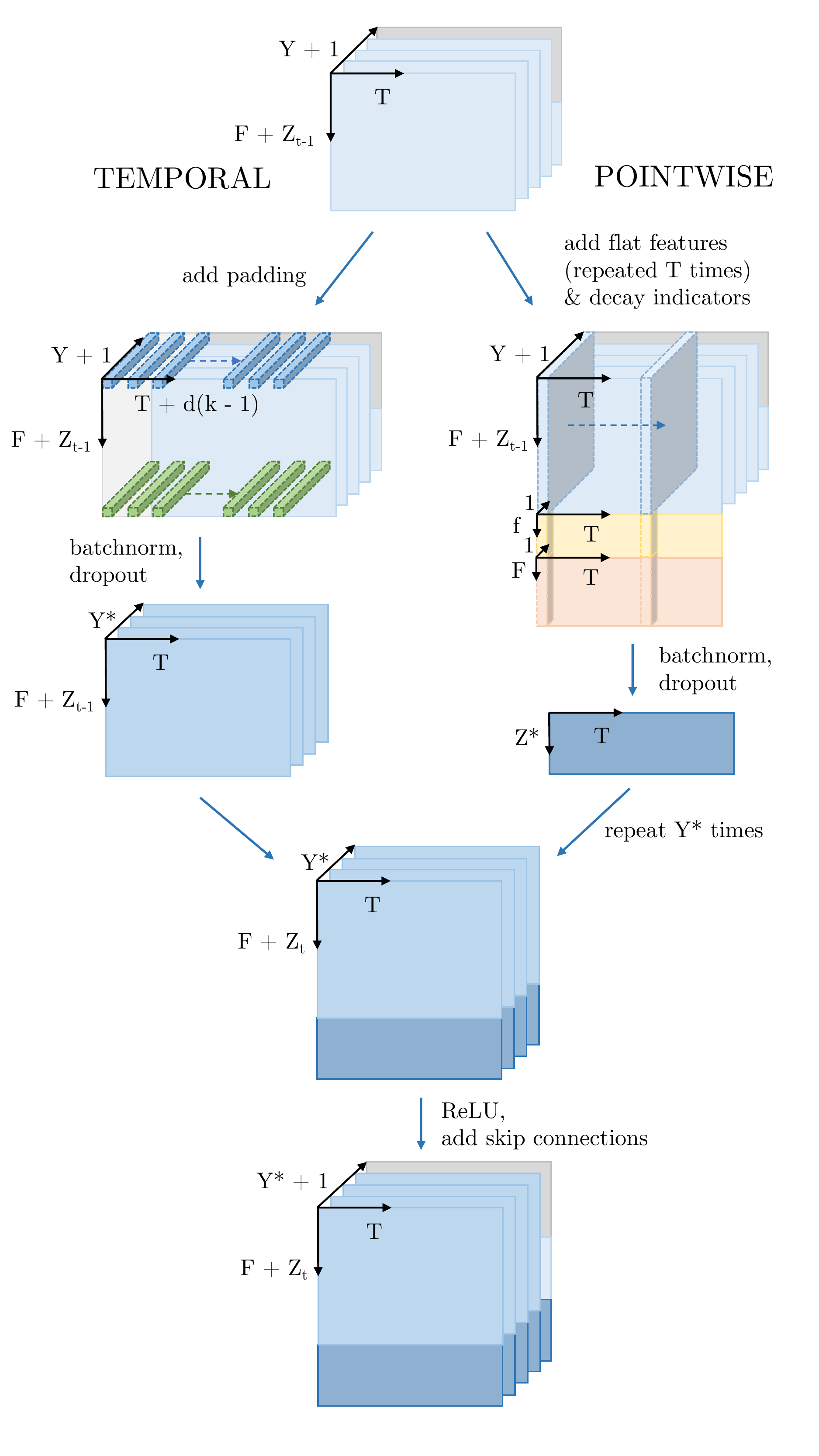}
  \caption{One layer of the TPC model. F is the number of time series features. T is the time series length. Y is the number of temporal channels per feature in the \textit{previous} TPC layer (except for the first layer where Y is 1; decay indicators (explained under `Time Series' in~\ref{timeseriespreproc}) make up this channel). Z$_{t-1}$ is the cumulative number of pointwise outputs from \textit{all previous} TPC layers. Y$^*$ and Z$^*$ are the number of temporal channels per feature and pointwise outputs respectively in the \textit{current} TPC layer. Z$_t =$ Z$_{t-1} +$ Z$^*$. The differently coloured temporal filters indicate independent parameters. $d$ is the temporal dilation, $k$ is the kernel size. Decay indicator features (\ref{preproc}) are shown in orange, $f$ flat features are shown in yellow. The skip connections consist of F original features (grey) and Z$_{t+1}$ pointwise outputs (light blue). We ignore the batch dimension for clarity.}
  \label{fig:TPC}
\end{figure}

\paragraph{Skip Connections}
\label{skip}
We propagate skip connections \citep{DBLP:journals/corr/HeZRS15} which allow each layer to see the original data and the pointwise outputs from previous layers. This helps the network to cope with infrequently sampled data (this is explored further in~\ref{appendix:skip}).

\paragraph{Temporal Pointwise Convolution}
Our model -- which we refer to as Temporal Pointwise Convolution (TPC) -- combines temporal and pointwise convolution in parallel. Figure~\ref{fig:TPC} shows just one TPC layer, however the full model has 9 TPC layers stacked sequentially. \ref{sec:modeloverview} contains further details on the surrounding (non-time series) architecture.

\paragraph{Loss Function}
The remaining LoS has a positive skew (\ref{rLOS}) which makes the task more challenging. We partly circumvent this by replacing the commonly used mean squared error (MSE) loss with mean squared \textit{log} error (MSLE). MSLE penalises \textit{proportional} over absolute error, which seems more reasonable when considering an error of 5 days in the context of a 2-day vs.\ a 30-day stay.

\begin{table*}[h]
  \vspace{-1.1em}
  \caption{Performance of the TPC model compared to the baselines (\textit{a}) and ablation studies (\textit{b}) and (\textit{c}). We report the mean absolute deviation (MAD), mean absolute percentage error (MAPE), mean squared error (MSE), mean squared log error (MSLE), coefficient of determination ($R^2$) and Cohen Kappa Score. The error margins are 95\% confidence intervals (CIs) calculated over 10 runs. Unless otherwise specified, the loss function is MSLE. (\textit{a}) shows the baseline comparisons. (\textit{b}) compares the effect of the loss function on the TPC model. (\textit{c}) shows ablations of the TPC model. WS refers to weight sharing between the features. The best results are highlighted in blue. If the result is statistically significant on a t-test then it is indicated with stars (* = p < 0.05, ** = p < 0.001).} 
  \label{tab:results}
  \centering
  \small
  \begin{minipage}{0.04\textwidth}
  \label{tab:baselineresults}
  \centering
  (a)
  \end{minipage}
  \begin{minipage}{0.95\textwidth}
  \centering
  \begin{tabular}{p{2.7cm}p{1.45cm}p{1.45cm}p{1.3cm}p{1.45cm}p{1.45cm}p{1.45cm}}
    \toprule
        \textbf{Model} & \textbf{MAD} & \textbf{MAPE} & \textbf{MSE} & \textbf{MSLE} & \boldmath{$R^2$} & \textbf{Kappa} \\
    \midrule
        Mean & \footnotesize{3.21} & \footnotesize{395.7} & \footnotesize{29.5} & \footnotesize{2.87} & \footnotesize{0.00} & \footnotesize{0.00} \\
        Median & \footnotesize{2.76} & \footnotesize{184.4} & \footnotesize{32.6} & \footnotesize{2.15} & \hspace{-0.32em}\footnotesize{-0.11} & \footnotesize{0.00} \\
        LSTM & \footnotesize{2.39$\pm$0.00} & \footnotesize{118.2$\pm$1.1} & \footnotesize{26.9$\pm$0.1} & \footnotesize{1.47$\pm$0.01} & \footnotesize{0.09$\pm$0.00} & \footnotesize{0.28$\pm$0.00} \\
        CW LSTM & \footnotesize{2.37$\pm$0.00} & \footnotesize{114.5$\pm$0.4} & \footnotesize{26.6$\pm$0.1} & \footnotesize{1.43$\pm$0.00} & \footnotesize{0.10$\pm$0.00} & \footnotesize{0.30$\pm$0.00} \\
        Transformer & \footnotesize{2.36$\pm$0.00} & \footnotesize{114.1$\pm$0.6} & \footnotesize{26.7$\pm$0.1} & \footnotesize{1.43$\pm$0.00} & \footnotesize{0.09$\pm$0.00} & \footnotesize{0.30$\pm$0.00} \\
        TPC & \footnotesize{\textBF{\textcolor{blue}{1.78$\pm$0.02}}}$^{\tiny{**}}$ & \footnotesize{\textBF{\textcolor{blue}{63.5$\pm$4.3}}}$^{\tiny{**}}$ & \footnotesize{\textBF{\textcolor{blue}{21.7$\pm$0.5}}}$^{\tiny{**}}$ & \footnotesize{\textBF{\textcolor{blue}{0.70$\pm$0.03}}}$^{\tiny{**}}$ & \footnotesize{\textBF{\textcolor{blue}{0.27$\pm$0.02}}}$^{\tiny{**}}$ & \footnotesize{\textBF{\textcolor{blue}{0.58$\pm$0.01}}}$^{\tiny{**}}$ \\
    \end{tabular}
    \end{minipage}
    \begin{minipage}{0.04\textwidth}
    \label{tab:lossfnresults}
    \centering
    (b)
    \end{minipage}
    \begin{minipage}{0.95\textwidth}
    \centering
    \begin{tabular}{p{2.7cm}p{1.45cm}p{1.45cm}p{1.3cm}p{1.45cm}p{1.45cm}p{1.45cm}}
    \toprule
        TPC (MSLE) & \footnotesize{\textBF{\textcolor{blue}{1.78$\pm$0.02}}}$^{\tiny{**}}$ & \footnotesize{\textBF{\textcolor{blue}{63.5$\pm$4.3}}}$^{\tiny{**}}$ & \footnotesize{21.7$\pm$0.5} & \footnotesize{\textBF{\textcolor{blue}{0.70$\pm$0.03}}}$^{\tiny{**}}$ & \footnotesize{\textBF{\textcolor{blue}{0.27$\pm$0.02}}} & \footnotesize{\textBF{\textcolor{blue}{0.58$\pm$0.01}}}$^{\tiny{**}}$ \\
        TPC (MSE) & \footnotesize{2.21$\pm$0.02} & \footnotesize{154.3$\pm$10.1} & \footnotesize{\textBF{\textcolor{blue}{21.6$\pm$0.2}}} & \footnotesize{1.80$\pm$0.10} & \footnotesize{\textBF{\textcolor{blue}{0.27$\pm$0.01}}} & \footnotesize{0.47$\pm$0.01} \\
    \end{tabular}
    \end{minipage}
    \begin{minipage}{0.04\textwidth}
    \label{tab:ablationresults}
    \centering
    (c)
    \end{minipage}
    \begin{minipage}{0.95\textwidth}
    \centering
    \begin{tabular}{p{2.7cm}p{1.45cm}p{1.45cm}p{1.3cm}p{1.45cm}p{1.45cm}p{1.45cm}}
    \toprule
        TPC & \footnotesize{\textBF{\textcolor{blue}{1.78$\pm$0.02}}}$^{\tiny{**}}$ & \footnotesize{\textBF{\textcolor{blue}{63.5$\pm$3.8}}}$^{\tiny{**}}$ & \footnotesize{\textBF{\textcolor{blue}{21.8$\pm$0.5}}}$^{\tiny{**}}$ & \footnotesize{\textBF{\textcolor{blue}{0.71$\pm$0.03}}}$^{\tiny{**}}$ & \footnotesize{\textBF{\textcolor{blue}{0.26$\pm$0.02}}}$^{\tiny{**}}$ & \footnotesize{\textBF{\textcolor{blue}{0.58$\pm$0.01}}}$^{\tiny{**}}$ \\
        Point.\ only & \footnotesize{2.68$\pm$0.15} & \footnotesize{137.8$\pm$16.4} & \footnotesize{29.8$\pm$2.9} & \footnotesize{1.60$\pm$0.03} & \hspace{-0.334em}\footnotesize{-0.01$\pm$0.10}\hspace{-0.334em} & \footnotesize{0.38$\pm$0.01} \\
        Temp.\ only & \footnotesize{1.91$\pm$0.01} & \footnotesize{71.2$\pm$1.1} & \footnotesize{23.1$\pm$0.2} & \footnotesize{0.86$\pm$0.01} & \footnotesize{0.22$\pm$0.01} & \footnotesize{0.52$\pm$0.01} \\
        Temp.\ only (WS) & \footnotesize{2.34$\pm$0.01} & \footnotesize{116.0$\pm$1.2} & \footnotesize{26.5$\pm$0.2} & \footnotesize{1.40$\pm$0.01} & \footnotesize{0.10$\pm$0.01} & \footnotesize{0.31$\pm$0.00} \\
    \bottomrule
    \end{tabular}
    \end{minipage}
\vspace{-0.3em}
\end{table*}

\section{Experiments and Results}
\paragraph{Data}
\label{data}
We use the eICU Database \citep{Pollard2018}, a multi-centre dataset from 208 hospitals in the United States. We extracted diagnoses, flat features and time series from all adult patients (\textgreater 18 years) with a LoS of at least 5 hours and at least one observation (more details in~\ref{cohort} and~\ref{preproc}). Our final cohort contained 118,534 unique patients and 146,666 ICU stays, which were split such that 70\%, 15\% and 15\% of patients were used for training, validation and testing respectively.

\paragraph{Baselines}
We include `mean' and `median' models that always predict 3.50 and 1.70 days respectively (the mean and median of the training set). Our standard LSTM baseline is very similar to \citet{sheikhalishahi2019benchmarking}. The channel-wise LSTM (CW LSTM) baseline consists of a set of independent LSTMs which process each feature separately (note the similarity with the TPC model). The Transformer is a multi-head self-attention model which, like TPC, is not constrained to progress one timestep at a time; however, unlike TPC, it is not able to scale its receptive fields or process features independently.

\paragraph{TPC Performance}
Table~\ref{tab:baselineresults}a shows that TPC outperforms all of the baseline models on every metric -- particularly those that are more robust to skewness: MAPE, MSLE and Kappa. The best performing \textit{baselines} are the Transformer and the channel-wise LSTM (CW LSTM). This is highly consistent with \citet{harutyunyan} (for CW LSTM) and \citet{2304ed73e858419398e3ee1508af5825} (for Transformers), who both found small improvements over LSTM.

\paragraph{MSLE Loss Function}
Table~\ref{tab:lossfnresults}b shows that using MSLE leads to significantly improved behaviour in the TPC model, with large performance gains in MAD, MAPE, MSLE and Kappa, while conceding little on MSE and $R^2$. The MSE results for the baseline models show a similar pattern (Table~\ref{tab:mseresults}).

\paragraph{Ablation Studies}
Table~\ref{tab:ablationresults}c shows ablations of the TPC model. The temporal-only model far outperforms the pointwise-only model, but neither is as successful as the complete TPC model. The temporal-only model is significantly and severely hindered by weight sharing, suggesting that having independent parameters per feature is important.

\section{Discussion}
\label{discussion}
We have shown that TPC outperforms all the baseline models on LoS. To explain its success, we start by examining its parallel architectures. Each component has been designed to extract different information: trends from the temporal arm and inter-feature relationships from the pointwise arm. The ablation study reveals that the temporal element is more important, but their contributions are complementary since the best performance is attained when they are used together.

Next, we highlight that the temporal-only model far outperforms its most direct comparison, the CW LSTM, on all metrics. In theory they are well matched because they both have feature-specific parameters but are restricted from learning cross-feature interactions. To explain this, we can consider how the information flows through the model. The temporal-only model can directly step across large time gaps, whereas the CW LSTM is forced to progress one timestep at a time. This gives the CW LSTM the harder task of remembering information across a noisy EHR with distracting signals of varying frequency. In addition, the temporal-only model can tune its receptive fields for optimal processing of each feature thanks to the skip connections (which are not present in the CW LSTM). 

The difference in performance between the temporal-only model with and without weight sharing highlights the importance of assigning independent parameters to each feature. EHR time series can be irregularly and sparsely sampled,  and they exhibit considerable variability in their temporal characteristics (evident in Figure~\ref{fig:one_patient}). This presents a challenge for any model, especially if it is constrained to learn one set of parameters to suit all features. 

Finally, we need to consider that \textit{periodicity} is a key property of EHR data (both in sampling patterns and biological functions e.g.\ sleep and medication schedules). The temporal component of the TPC model has an inherent periodic structure which makes it much easier to learn EHR trends. By comparison, a single attention head in the Transformer does not look at timepoints a fixed distance apart, but can take an arbitrary form. This is helpful in NLP but not for processing EHRs.

Regarding the choice of loss function, we reiterate that using MSLE greatly mitigates for positive skew in the LoS task, and the benefit is not model-specific. This demonstrates that careful consideration of the task -- as well as the data and model -- is an important step towards building useful tools in healthcare.


\section*{Acknowledgements}
The authors would like to thank Alex Campbell, Petar Veli\v{c}kovi\'{c}, and Ari Ercole for helpful discussions and advice. We would also like to thank Louis-Pascal Xhonneux, C\u{a}t\u{a}lina Cangea and Nikola Simidjievski for their help in reviewing the manuscript. Finally we thank the Armstrong Fund, the Frank Edward Elmore Fund, and the School of Clinical Medicine at the University of Cambridge for their generous funding.

\bibliography{references}

\newpage
\newpage
\onecolumn
\appendix

\section{Implementation Details}
\subsection{Model Architecture}
\label{sec:modeloverview}

\begin{figure}[h]
  \centering
  \hspace{7em}
  \includegraphics[scale=0.5]{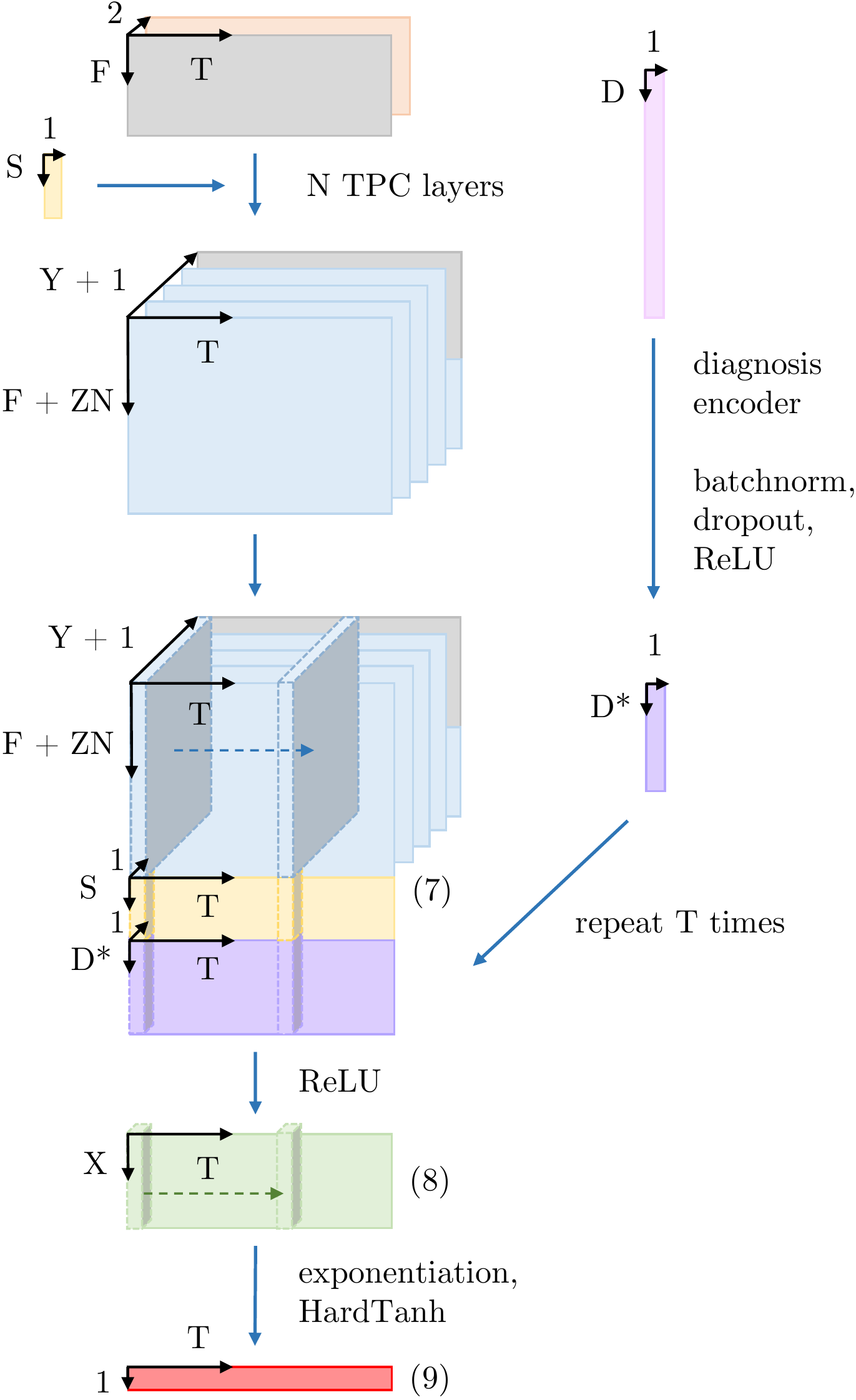}
  \caption{Overview of the framework used for the TPC model and the baselines. F, T, Y$^*$, Z$^*$, Z$_t$ and $f$ are defined in the caption to Figure~\ref{fig:TPC}. The original time series (grey) along with the decay indicators (orange) (explained under `Time Series' in Section~\ref{timeseriespreproc}) are processed by $n$ TPC layers. If a baseline model were used instead of TPC, the time series output dimensions would be M x T, where M is the LSTM hidden size or $d_{model}$ in the Transformer (this is in place of the light blue and grey output in the TPC model). The diagnoses, $d$, are embedded by a diagnosis encoder -- a single fully connected layer of size D. The time series (blue and grey), diagnosis embedding (purple) and flat features (yellow) are concatenated along the feature axis, and a two-layer pointwise convolution is applied to obtain the final predictions (red).}
    \label{fig:modeloverview}
\end{figure}

Figure~\ref{fig:modeloverview} shows the full architecture. With each successive TPC layer, we increase the temporal dilation by 1. After processing the time series, we combine them with flat (non time-varying) features and a diagnosis embedding (Figure~\ref{fig:modeloverview} shows the complete architecture). The combined features pass through a small two-layer pointwise convolution to obtain the LoS predictions. We apply an exponential function before the predictions are returned. This is intended to help to circumvent a common issue seen in previous models (e.g.\ \citet{harutyunyan}, as they struggle to produce predictions over the full range of length of stays). It effectively allows the upstream network to model $\log(\text{LoS})$ instead of LoS. This distribution is much closer to a Gaussian distribution than the remaining LoS distribution. We use batch normalisation \citep{Ioffe2015} and dropout \citep{Srivastava2014} to regularise the model. The hyperparameter search methodology is described in Section~\ref{hyperparamsearch}. Finally, we apply a HardTanh function \citep{DBLP:journals/corr/GulcehreMDB16} to the output to clip any predictions that are smaller than 30 minutes or larger than 100 days, which protects against inflated loss values:
\begin{equation*}
        \text{HardTanh}(x) =
        \begin{cases}
            100,          & \text{if } x~\text{\textgreater}~100,\\
            \frac{1}{48},          & \text{if } x~\text{\textless}~\frac{1}{48},\\
            x,         & \text{otherwise.}
        \end{cases}
\end{equation*}

\subsection{Remaining Length of Stay Task}

\label{rLOS}
\begin{figure}[h]
  \centering
  \includegraphics[width=0.5\textwidth]{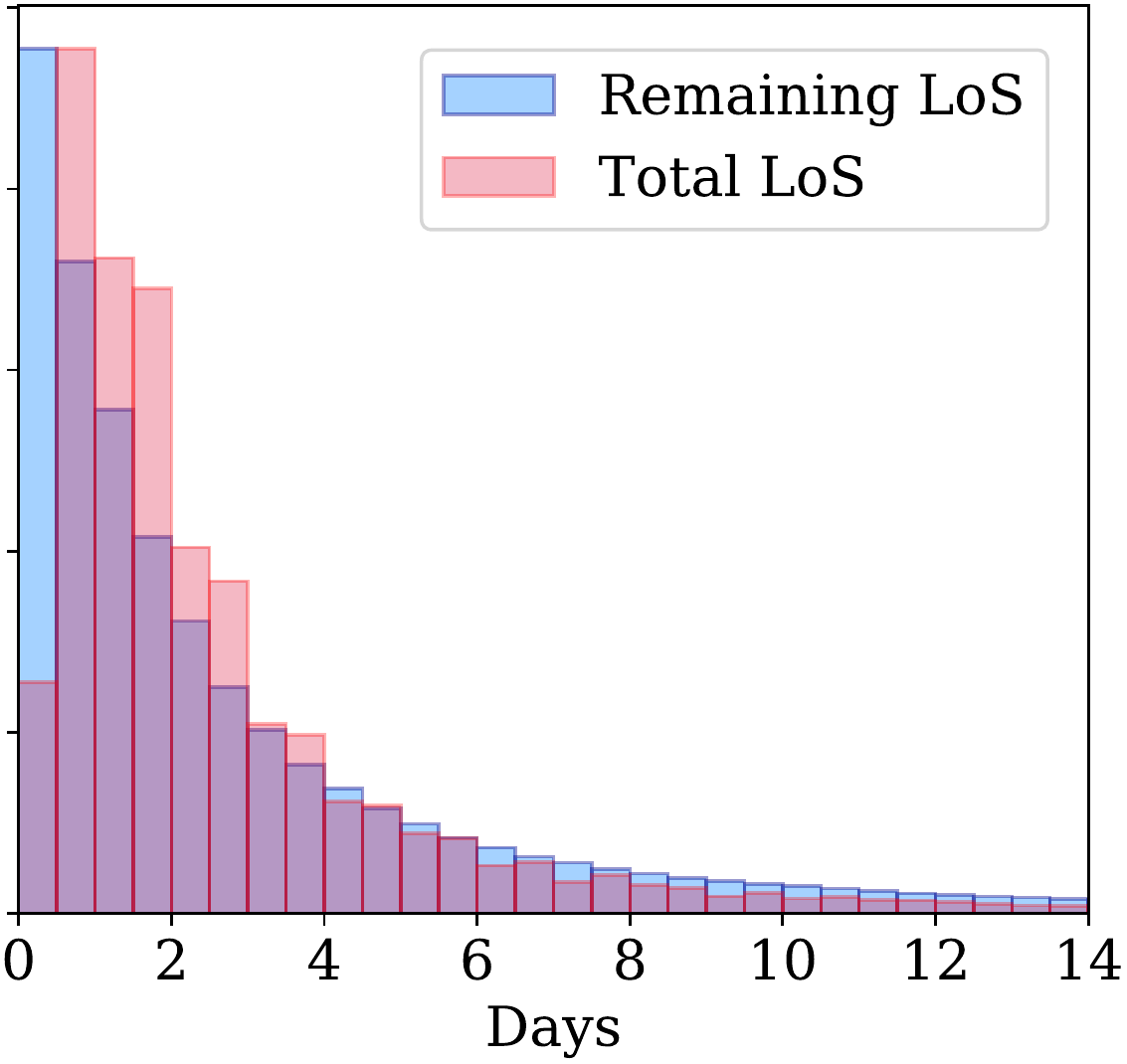}
  \caption{Total and remaining length of stay distributions. The mean and median values are (2.99, 1.82) and (3.47, 1.67) days respectively.}
\label{fig:los_dist}
\end{figure}

We assign a remaining LoS target to each hour of the stay, beginning 5 hours after admission to the ICU and ending when the patient dies or is discharged. The remaining LoS is calculated by subtracting the time elapsed in the ICU from the total LoS. We only train on data within the first 14 days of any patient's stay to protect against batches becoming overly long and slowing down training. This cut-off applies to 2.4\% of patient stays, but it does \textit{not} affect their maximum remaining length of stay values because these appear within the first 14 days. The remaining LoS distribution is shown in Figure~\ref{fig:los_dist}.

\subsection{Feature Selection}
\label{cohort}
We selected time series variables from the following tables: \textit{lab}, \textit{nursecharting}, \textit{respiratorycharting}, \textit{vitalperiodic} and \textit{vitalaperiodic}. We used a semi-automatic process for feature selection. To be included, the variable had to be present in at least 12.5\% of patient stays, or 25\% of stays for \textit{lab} variables. The \textit{lab} table contains a much larger number of variables than the other tables, and they tend to be sparsely sampled (once per day or less). We extracted diagnoses from the \textit{pasthistory}, \textit{admissiondx} and \textit{diagnoses} tables, and 17 flat (non time-varying) features from the \textit{patient}, \textit{apachepatientresult} and \textit{hospital} tables (see Table~\ref{tab:flat}). 

\begin{table}[h]
    \caption{Flat features used in the model. Age \textgreater 89, Null Height and Null Weight were added as indicator variables to indicate when the age was more than 89 but has been capped, and when the height or weight were missing and have been imputed with the mean value.}
    \label{tab:flat}
    \centering
    \begin{tabular}{lll}
        \toprule
        \textbf{Feature} & \textbf{Type} & \textbf{Source Table} \\
        \midrule
        Gender & Binary & patient \\
        Age & Discrete & patient \\
        Hour of Admission & Discrete & patient \\
        Height & Continuous & patient \\
        Weight & Continuous & patient \\
        Ethnicity & Categorical & patient \\
        Unit Type & Categorical & patient \\
        Unit Admit Source & Categorical & patient \\
        Unit Visit Number & Categorical & patient \\
        Unit Stay Type & Categorical & patient \\
        Num Beds Category & Categorical & hospital \\
        Region & Categorical & hospital \\
        Teaching Status & Binary & hospital \\
        Physician Speciality & Categorical & apachepatientresult \\
        Age \textgreater 89 & Binary & \\
        Null Height & Binary & \\
        Null Weight & Binary & \\
        \bottomrule
    \end{tabular}
\end{table}

\subsection{Feature Pre-processing}
\label{preproc}
\paragraph{Flat Features}
Discrete and continuous variables were scaled to the interval [-1, 1], using the 5th and 95th percentiles as the boundaries, and absolute cut offs were placed at [-4, 4]. This was to protect against large or erroneous inputs, while avoiding assumptions about the variable distributions. Binary variables were coded as 1 and 0. Categorical variables were converted to one-hot encodings.

\begin{figure}[h]
  \centering
  \includegraphics[scale=0.5]{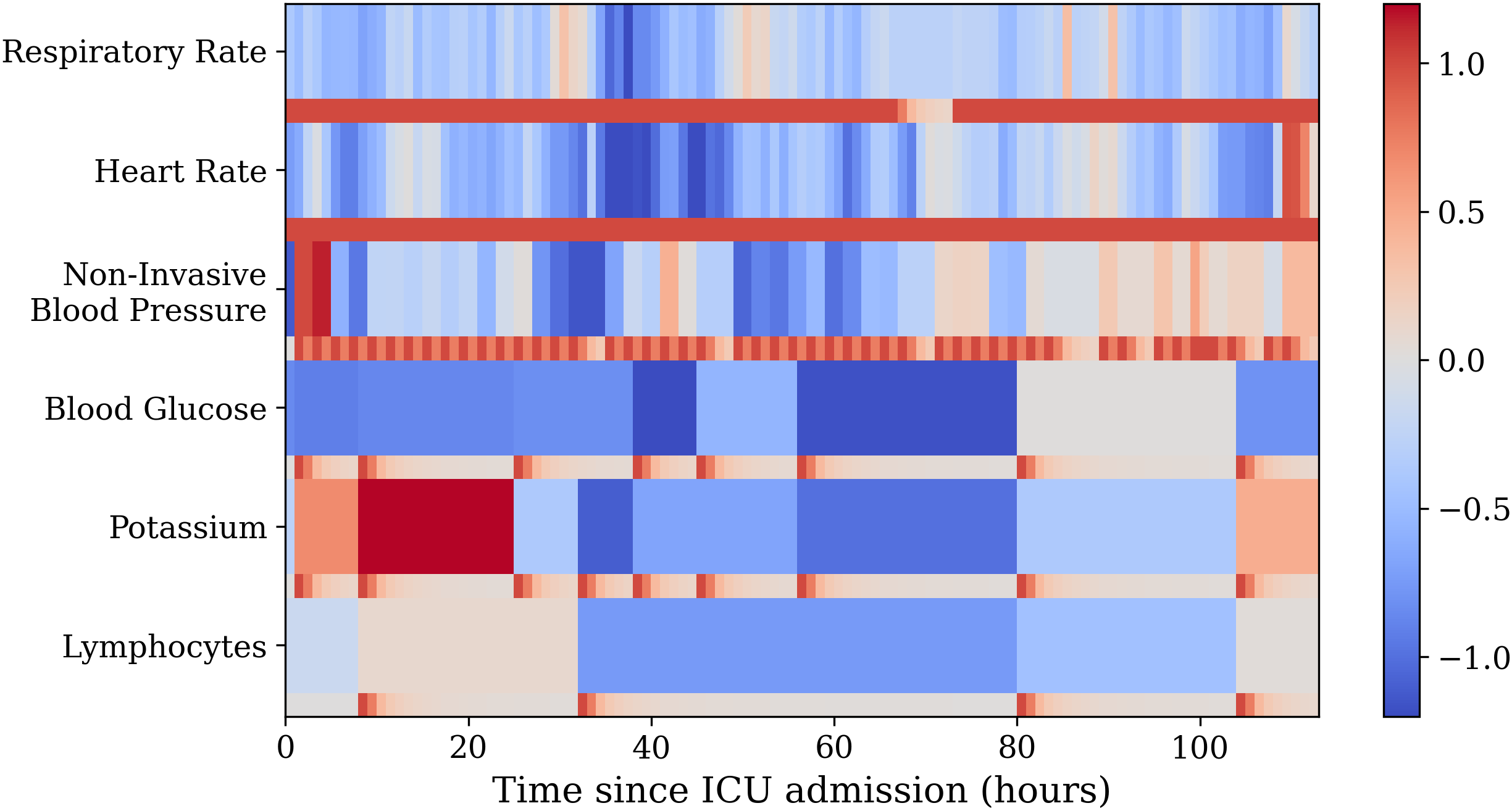}
  \caption{A selection of variables from one patient (after pre-processing). The colour scale indicates the value of the feature. Each feature is accompanied by its decay indicator (narrow bar underneath). Blood Glucose, Potassium and Lymphocytes are all laboratory tests. These are performed approximately once per day but we can see that the frequency is irregular. Non-invasive blood pressure is a variable that is recorded by the nurse. The sampling frequency is around 2 hours and is quite consistent. Respiratory Rate and Heart Rate are vital signs that are automatically logged at regular intervals.}
  \label{fig:one_patient}
\end{figure}

\paragraph{Time Series}
\label{timeseriespreproc}
For each admission, 87 time-varying features (Table~\ref{tab:timeseries}) were extracted from each hour of the ICU visit, and up to 24 hours before the ICU visit. The variables were processed in the same manner as the flat features. In general, the sampling is very irregular, so the data was re-sampled according to one hour intervals. Where there is missingness, we forward-fill to bridge the gaps in the data\footnote{This is preferable to interpolating between the data points because in realistic scenarios the clinician would only have the most recent value and its timestamp.}. After forward-filling is complete, any data recorded before the ICU admission is removed. To inform the model about where the data is stale, we add a `decay indicator' to each feature to track how long it has been since a genuine observation was recorded. The decay is calculated as $0.75^t$, where $t$ is the time since the last recording. If it is up to date, decay is 1, and if it cannot be forward-filled, decay is 0. This is similar in spirit to the masking used by \citet{Che2018}. A real example is shown in Figure~\ref{fig:one_patient}.

\begin{table*}[h]
    \caption{Time Series features. `Time in the ICU' and `Time of day' were not part of the tables in eICU but were added later as helpful indicators to the model.}
    \label{tab:timeseries}
    \small
    \centering
    \begin{tabular}{llll}
        \toprule
        \multicolumn{4}{c}{\textbf{Source Table}} \\
        \multicolumn{3}{l}{\textbf{\textit{lab}}} & \textbf{\textit{respiratorycharting}} \\
        \midrule
        -basos & MPV & glucose & Exhaled MV \\
        -eos & O2 Sat (\%) & lactate & Exhaled TV (patient) \\
        -lymphs & PT & magnesium & LPM O2 \\
        -monos & PT - INR & pH & Mean Airway Pressure \\
        -polys & PTT & paCO2 & Peak Insp. Pressure \\
        ALT (SGPT) & RBC & paO2 & PEEP \\
        AST (SGOT) & RDW & phosphate & Plateau Pressure \\
        BUN & WBC x 1000 & platelets x 1000 & Pressure Support \\
        Base Excess & albumin & potassium & RR (patient) \\
        FiO2 & alkaline phos. & sodium & SaO2 \\
        HCO3 & anion gap & total bilirubin & TV/kg IBW \\
        Hct & bedside glucose & total protein & Tidal Volume (set) \\
        Hgb & bicarbonate & troponin - I & Total RR \\
        MCH & calcium & urinary specific gravity & Vent Rate \\
        MCHC & chloride & & \\
        MCV & creatinine & & \\
        \vspace{-0.8em}\\
        \toprule
        \textbf{\textit{nursecharting}} &  \textbf{\textit{vitalperiodic}} & \textbf{\textit{vitalaperiodic}} & \textbf{N/A} \\
        \midrule
        Bedside Glucose & cvp & noninvasivediastolic & Time in the ICU \\
        Delirium Scale/Score & heartrate & noninvasivemean & Time of day \\
        Glasgow coma score & respiration & noninvasivesystolic & \\
        Heart Rate & sao2 & & \\
        Invasive BP & st1 & & \\
        Non-Invasive BP & st2 & & \\
        O2 Admin Device & st3 & & \\
        O2 L/\% & systemicdiastolic & & \\
        O2 Saturation & systemicmean & & \\
        Pain Score/Goal & systemicsystolic & & \\
        Respiratory Rate & temperature & & \\
        Sedation Score/Goal & & & \\
        Temperature & & & \\
        \bottomrule
    \end{tabular}
\end{table*}

\paragraph{Diagnoses}
Like many EHRs, diagnosis coding in eICU is hierarchical. At the lowest level they can be quite specific e.g.\ ``neurologic $|$ disorders of vasculature $|$ stroke $|$ hemorrhagic stroke $|$ subarachnoid hemorrhage $|$ with vasospasm''. To maintain the hierarchical structure within a flat vector, we assigned separate features to each hierarchical level and use binary encoding. This produces a vector of size 4,436 with an average sparsity of 99.5\% (only 0.5\% of the data is positive). We apply a 1\% prevalence cut-off on all these features to reduce the size of the vector to 293 and the average sparsity to 93.3\%. If a disease does not make the cut-off for inclusion, it is still included via any parent classes that do make the cut-off (in the above example we record everything up to ``subarachnoid hemorrhage''). We only included diagnoses that were recorded before the 5th hour in the ICU, to avoid leakage from the future.

Many diagnostic and interventional coding systems are hierarchical in nature: ICD-10 classification \citep{icd10}, Clinical Classifications Software \citep{ccs}, SNOMED CT \citep{10.1016/j.cmpb.2011.01.002} and OPCS Classification of Interventions and Procedures \citep{Digital2019}, so this technique is generalisable to other coding systems present in EHRs. 

\newpage
\subsection{Baselines}
\label{baselines}
A certain level of performance is achievable `for free' just by predicting values that are close to the mean or median LoS (3.50 and 1.70 days respectively). We include these static models as elemental baselines. Our `stronger' baselines are a two-layer \textit{Long-Short Term Memory} network (LSTM) \citep{10.1162/neco.1997.9.8.1735}, a two-layer channel-wise LSTM, and a Transformer encoder network \citep{46201} (see Section~\ref{hyperparamsearch} for their hyperparameters). 

Our standard LSTM baseline is very similar to the one used in a recent eICU benchmark paper including LoS prediction \citep{sheikhalishahi2019benchmarking}. The channel-wise LSTM (CW LSTM) consists of a set of independent LSTMs that process each feature separately. Like our TPC model, the CW LSTM has independent parameters dedicated to each time series feature. In theory, it should be able to cope better with irregular sampling and varying frequencies in the data, but it may be hindered by the inability to compute cross-feature interactions along the way. \citet{harutyunyan} found that they performed better than the standard LSTM.

The Transformer was shown to perform marginally better than the standard LSTM when predicting LoS \citep{2304ed73e858419398e3ee1508af5825}. It is a multi-head self-attention model, originally designed for sequence-to-sequence tasks in natural language processing. It consists of both an encoder and decoder, however we only use the former because the LoS task is regression. Our implementation is the same as the original encoder in \citet{46201}, except that we add temporal masking to impose causality\footnote{The processing of each timepoint can only depend on current or earlier positions in the sequence}, and we omit the positional encodings because they were not helpful for the LoS task (see Table~\ref{tab:Transformerhyperparams}). Hypothetically, the Transformer shares an advantage with TPC in that it is not constrained to progress one timestep at a time. However, it is not able to scale its receptive fields in the same way as TPC and it does not have independent parameters per feature. 

In all network baselines, the data (including decay indicators) and the non-time series components of the models were the same as in TPC (Figure~\ref{fig:modeloverview}).

\subsection{Hyperparameter Search}
\label{hyperparamsearch}
Our model and its baselines have hyperparameters that can broadly be split into three categories: time series specific, non-time series specific and global parameters (shown in more detail in Tables~\ref{tab:TPChyperparams}, \ref{tab:LSTMhyperparams} and \ref{tab:Transformerhyperparams}). The hyperparameter search ranges have been included in Table~\ref{tab:hyperparamsearch}. First, we ran 25 randomly sampled hyperparameter trials on the TPC model to decide the non-time series specific parameters (diagnosis embedding size, final fully connected layer size, batch normalisation strategy and dropout rate) keeping all other parameters fixed. These parameters then remained fixed for all the models which share their non-time series specific architecture. We ran 50 hyperparameter trials to optimise the remaining parameters for the TPC, standard LSTM, and Transformer models. To train the channel-wise LSTM and the temporal model with weight sharing, we ran a further 10 trials to re-optimise the hidden size (8 per feature) and number of temporal channels (32 channels shared across all features) respectively. For all other ablation studies and variations of each model, we kept the same hyperparameters where applicable (see Table~\ref{tab:results} for a full list of all the models). The number of epochs was determined by selecting the best validation performance from a model trained over 50 epochs. This was different for each model: 8 for LSTM, 30 for CW LSTM, and 15 for the Transformer and TPC models. We noted that the best LSTM hyperparameters were similar to that found in \citet{sheikhalishahi2019benchmarking}. We used trixi to structure our experiments and easily compare different hyperparameter choices \citep{trixi2017}. All deep learning methods were implemented in PyTorch \citep{NEURIPS2019_9015} and were optimised using Adam \citep{KingmaB14}. 

\begin{table*}[h]
  \caption{The TPC model has 11 hyperparameters (Main Dropout and Batch Normalisation have been repeated in the table because they apply to multiple parts of the model). We allowed the model to optimise a custom dropout rate for the temporal convolutions because they have fewer parameters and might need less regularisation than the rest of the model. The best hyperparameter values are shown in brackets. Hyperparameters marked with * were fixed across all of the models.}
  \label{tab:TPChyperparams}
  \centering
  \begin{tabular}{ll}
    \toprule
    \multicolumn{2}{c}{\textbf{TPC Specific}}\\
    \textbf{Temporal Specific} & \textbf{Pointwise Specific}\\
    \midrule
    Temp.\ Channels (12)&Point.\ Channels (13)\\
    Temp.\ Dropout (0.05)&Main Dropout* (0.45)\\
    Kernel Size (4)& \\
    \multicolumn{2}{c}{Batch Normalisation* (True)}\\
    \multicolumn{2}{c}{No. TPC Layers (9)}\\
    \vspace{-0.8em}\\
    \toprule
    \textbf{Non-TPC Specific} & \textbf{Global Parameters}\\
    \midrule
    Diag. Embedding Size* (64)&Batch Size (32)\\
    Main Dropout* (0.45)&Learning Rate (0.00226)\\
    Final FC Layer Size* (17)&\\
    Batch Normalisation* (True)&\\
    \bottomrule
  \end{tabular}
\end{table*}

\begin{table*}[h]
  \caption{The LSTM model has 9 hyperparameters. We allowed the model to optimise a custom dropout rate for the LSTM layers. Note that batch normalisation is not applicable to the LSTM layers. The best hyperparameter values are shown in brackets. Hyperparameters marked with * were fixed across all of the models.}
  \label{tab:LSTMhyperparams}
  \centering
  \begin{tabular}{lll}
    \toprule
    \textbf{LSTM Specific} & \textbf{Non-LSTM Specific} & \textbf{Global Parameters}\\
    \midrule
    Hidden State (128)&Diag. Embedding Size* (64)&Batch Size (512)\\
    LSTM Dropout (0.2)&Main Dropout* (0.45)&Learning Rate (0.00129)\\
    No. LSTM Layers (2)&Final FC Layer Size* (17)&\\
    &Batch Normalisation* (True)&\\
    \bottomrule
  \end{tabular}
\end{table*}

\begin{table*}[h]
  \caption{The Transformer model has 12 hyperparameters. We allowed the model to optimise a custom dropout rate for the Transformer layers. The positional encoding hyperparameter is binary; it determines whether or not we used the original positional encodings proposed by \citet{46201}. They were not found to be helpful (perhaps because we already have a feature to indicate the position in the time series (Section~\ref{timeseriespreproc})). Note that batch normalisation is not applicable to the Transformer layers (the default implementation uses layer normalisation). The best hyperparameter values are shown in brackets. Hyperparameters marked with * were fixed across all of the models.}
  \label{tab:Transformerhyperparams}
  \centering
  \begin{tabular}{lll}
    \toprule
    \textbf{Transformer Specific} & \textbf{Non-Transformer Specific} & \textbf{Global Parameters} \\
    \midrule
    No. Attention Heads (2)&Diag. Embedding Size* (64)&Batch Size (32)\\
    Feedforward Size (256)&Main Dropout* (0.45)&Learning Rate (0.00017)\\
    $d_{model}$ (16)&Final FC Layer Size* (17)&\\
    Positional Encoding (False)&Batch Normalisation* (True)&\\
    Transformer Dropout (0)&&\\
    No. Transformer Layers (6)&&\\
    \bottomrule
  \end{tabular}
\end{table*}

\newpage

\begin{table*}[h]
  \caption{Hyperparameter Search Ranges. We took a random sample from each range and converted to an integer if necessary. For the kernel sizes (not shown in the table) the range was dependent on the number of TPC layers selected (because large kernel sizes combined with a large number of layers can have an inappropriately wide range as the dilation factor increases per layer). In general the range of kernel sizes was around 2-5 (but it could be up to 10 for small numbers of TPC Layers).}
  \label{tab:hyperparamsearch}
  \centering
  \begin{tabular}{llll}
    \toprule
    \textbf{Hyperparameter} & \textbf{Lower} & \textbf{Upper} & \textbf{Scale}\\
    \midrule
    Batch Size & 4 & 512 & $\log_2$\\
    Dropout Rate (all) & 0 & 0.5 & Linear\\
    Learning Rate & 0.0001 & 0.01 & $\log_{10}$\\
    Batch Normalisation & True & False & \\
    Positional Encoding & True & False & \\
    Diagnosis Embedding Size & 16 & 64 & $\log_2$\\
    Final FC Layer Size & 16 & 64 & $\log_2$\\
    Channel-Wise LSTM Hidden State Size & 4 & 16 & $\log_2$\\
    Point.\ Channels & 4 & 16 & $\log_2$\\
    Temp.\ Channels & 4 & 16 & $\log_2$\\
    Temp.\ Channels (weight sharing) & 16 & 64 & $\log_2$\\
    LSTM Hidden State Size & 16 & 256 & $\log_2$\\
    $d_{model}$ & 16 & 256 & $\log_2$\\
    Feedforward Size & 16 & 256 & $\log_2$\\
    No. Attention Heads & 2 & 16 & $\log_2$\\
    No. TPC Layers & 1 & 12 & Linear\\
    No. LSTM Layers & 1 & 4 & Linear\\
    No. Transformer Layers & 1 & 10 & Linear\\
    \bottomrule
  \end{tabular}
\end{table*}

\FloatBarrier

\section{Evaluation Metrics}
\label{evaluationmetrics}
The metrics we use are: mean absolute deviation (MAD), mean absolute percentage error (MAPE), mean squared error (MSE), mean squared loss error (MSLE), coefficient of determination ($R^2$) and Cohen Kappa Score. We use 6 different metrics because there is a risk that bad models can `cheat' particular metrics just by virtue of being close to the mean or median value, or by not predicting long length of stays. This is not what we want in a bed management model, because long length of stays are disproportionately important due to their lasting effect on occupancy.

We have modified the MAPE metric slightly so that very small true LoS values do not produce unbounded MAPE values. We place a 4 hour lower bound on the divisor i.e. 
\begin{equation*}
    \text{Absolute Percentage Error} = \abs{\frac{y_{true} - y_{pred}}{\max{(y_{true}, \frac{4}{24})}}}*100
\end{equation*} 
MAD and MAPE are improved by centering predictions on the median. Likewise, MSE and $R^2$ are bettered by centering predictions around the mean. They are more affected by the skew. MSLE is a good metric for this task, indeed, it is the loss function in most experiments, but is less readily-interpretable than some of the other measures. Cohen's linear weighted Kappa Score \citep{doi:10.1177/001316446002000104} is intended for ordered classification tasks rather than regression, but it can effectively mitigate for skew if the bins are chosen well. It has previously provided useful insights in \citet{harutyunyan}, so we use the same LoS bins: 0-1, 1-2, 2-3, 3-4, 4-5, 5-6, 6-7, 7-8, 8-14, and 14+ days. As a classification measure, it will treat everything falling within the same classification bin as equal, so it is fundamentally a coarser measure than the other metrics.

To illustrate the importance of using multiple metrics, consider that the mean and median models are in some sense equally poor (neither has learned anything meaningful for our purposes). Nevertheless, the median model is able to better exploit the MAD, MAPE and MSLE metrics, and the mean model fares better with MSE, but the Kappa score betrays them both. A good model will perform well across all of the metrics.

\newpage
\section{Additional Investigations}
\subsection{Loss Function}

\begin{table*}[h]
    \caption{The effect of training with the mean squared logarithmic error (MSLE) loss function when compared to mean squared error (MSE). The metrics are defined in Section~\ref{evaluationmetrics}. The colour scheme and confidence interval calculation is described in the legend to Table~\ref{tab:results}.}
    \label{tab:mseresults}
    \centering
    \begin{tabular}{lllllll}
    \toprule
        \textbf{Model} & \textbf{MAD} & \textbf{MAPE} & \textbf{MSE} & \textbf{MSLE} & \boldmath{$R^2$} & \textbf{Kappa} \\
    \midrule
        LSTM (MSLE) & \footnotesize{\textBF{\textcolor{blue}{2.39$\pm$0.00}}} & \footnotesize{\textBF{\textcolor{blue}{118.2$\pm$1.1}}} & \footnotesize{26.9$\pm$0.1} & \footnotesize{\textBF{\textcolor{blue}{1.47$\pm$0.01}}} & \footnotesize{0.09$\pm$0.00} & \footnotesize{\textBF{\textcolor{blue}{0.28$\pm$0.00}}} \\
        LSTM (MSE) & \footnotesize{2.57$\pm$0.03} & \footnotesize{235.2$\pm$6.2} & \footnotesize{\textBF{\textcolor{blue}{24.5$\pm$0.2}}} & \footnotesize{1.97$\pm$0.02} & \footnotesize{\textBF{\textcolor{blue}{0.17$\pm$0.01}}} & \footnotesize{\textBF{\textcolor{blue}{0.28$\pm$0.01}}} \\
    \midrule
        CW LSTM (MSLE) & \footnotesize{\textBF{\textcolor{blue}{2.37$\pm$0.00}}} & \footnotesize{\textBF{\textcolor{blue}{114.5$\pm$0.4}}} & \footnotesize{26.6$\pm$0.1} & \footnotesize{\textBF{\textcolor{blue}{1.43$\pm$0.00}}} & \footnotesize{0.10$\pm$0.00} & \footnotesize{0.30$\pm$0.00} \\
        CW LSTM (MSE) & \footnotesize{2.56$\pm$0.01} & \footnotesize{218.5$\pm$4.0} & \footnotesize{\textBF{\textcolor{blue}{24.2$\pm$0.1}}} & \footnotesize{1.84$\pm$0.02} & \footnotesize{\textBF{\textcolor{blue}{0.18$\pm$0.00}}} & \footnotesize{\textBF{\textcolor{blue}{0.34$\pm$0.01}}} \\
    \midrule
        Transformer (MSLE) & \footnotesize{\textBF{\textcolor{blue}{2.36$\pm$0.00}}} & \footnotesize{\textBF{\textcolor{blue}{114.1$\pm$0.6}}} & \footnotesize{26.7$\pm$0.1} & \footnotesize{\textBF{\textcolor{blue}{1.43$\pm$0.00}}} & \footnotesize{0.09$\pm$0.00} & \footnotesize{\textBF{\textcolor{blue}{0.30$\pm$0.00}}} \\
        Transformer (MSE) & \footnotesize{2.51$\pm$0.01} & \footnotesize{212.7$\pm$5.2} & \footnotesize{\textBF{\textcolor{blue}{24.7$\pm$0.2}}} & \footnotesize{1.87$\pm$0.03} & \footnotesize{\textBF{\textcolor{blue}{0.16$\pm$0.01}}} & \footnotesize{0.28$\pm$0.01} \\
    \midrule
        TPC (MSLE) & \footnotesize{\textBF{\textcolor{blue}{1.78$\pm$0.02}}} & \footnotesize{\textBF{\textcolor{blue}{63.5$\pm$4.3}}} & 
        \footnotesize{21.7$\pm$0.5} & \footnotesize{\textBF{\textcolor{blue}{0.70$\pm$0.03}}} & \footnotesize{\textBF{\textcolor{blue}{0.27$\pm$0.02}}} & \footnotesize{\textBF{\textcolor{blue}{0.58$\pm$0.01}}} \\
        TPC (MSE) & \footnotesize{2.21$\pm$0.02} & \footnotesize{154.3$\pm$10.1} & \footnotesize{\textBF{\textcolor{blue}{21.6$\pm$0.2}}} & \footnotesize{1.80$\pm$0.10} & \footnotesize{\textBF{\textcolor{blue}{0.27$\pm$0.01}}} & \footnotesize{0.47$\pm$0.01} \\
    \bottomrule
    \end{tabular}
\end{table*}

\subsection{Skip Connections}
\label{appendix:skip}
We propagate skip connections \citep{DBLP:journals/corr/HeZRS15} to allow each layer to see the original data and the pointwise outputs from previous layers. This helps the network to cope with sparsely sampled data. For example, suppose a particular blood test is taken once per day. In order to not to lose temporal resolution, we forward-fill these data (\ref{cohort}) and convolve with increasingly dilated temporal filters until we find the appropriate width to capture a useful trend. However, if the smaller filters in previous layers (which did not see any useful trend) have polluted the original data by re-weighting, learning will be harder. Skip connections provide a consistent anchor to the input. They are concatenated (like in DenseNet \citep{densenet}, and are arranged in the shared-source connection formation \citep{wang20188}). The skip connections expand both the feature dimension to accommodate the pointwise outputs, and the channel dimension to fit the original data. This is best visualised in Figure~\ref{fig:TPC}. We can see from Table~\ref{tab:skipresults} that removing the skip connections reduces performance by 5-25\%.

\begin{table*}[h]
    \caption{The effect of skip connections. The metrics are defined in Section~\ref{evaluationmetrics}. The colour scheme and confidence interval calculation is described in the legend to Table~\ref{tab:results}.}
    \label{tab:skipresults}
    \centering
    \begin{tabular}{lllllll}
    \toprule
        \textbf{Model} & \textbf{MAD} & \textbf{MAPE} & \textbf{MSE} & \textbf{MSLE} & \boldmath{$R^2$} & \textbf{Kappa} \\
    \midrule
        TPC & \footnotesize{\textBF{\textcolor{blue}{1.78$\pm$0.02}}}$^{\tiny{**}}$ & \footnotesize{\textBF{\textcolor{blue}{63.5$\pm$3.8}}}$^{\tiny{**}}$ & \footnotesize{\textBF{\textcolor{blue}{21.8$\pm$0.5}}}$^{\tiny{**}}$ & \footnotesize{\textBF{\textcolor{blue}{0.71$\pm$0.03}}}$^{\tiny{**}}$ & \footnotesize{\textBF{\textcolor{blue}{0.26$\pm$0.02}}}$^{\tiny{**}}$ & \footnotesize{\textBF{\textcolor{blue}{0.58$\pm$0.01}}}$^{\tiny{**}}$ \\
        TPC (no skip) & \footnotesize{1.93$\pm$0.01} & \footnotesize{73.9$\pm$1.9} & \footnotesize{23.0$\pm$0.2} & \footnotesize{0.89$\pm$0.01} & \footnotesize{0.22$\pm$0.01} & \footnotesize{0.51$\pm$0.01} \\
    \bottomrule
    \end{tabular}
\end{table*}
\end{document}